\documentclass[letterpaper, 10 pt, journal,twoside]{IEEEtran}  

\IEEEoverridecommandlockouts                              

\usepackage{multirow}
\usepackage[noadjust]{cite}
\usepackage{graphics} 
\usepackage{epsfig} 
\usepackage{times} 
\usepackage{amsmath} 
\usepackage{amssymb}  
\usepackage{url}
\usepackage{gensymb}
\usepackage{graphicx}
\usepackage{amsmath}
\usepackage{booktabs}
\usepackage{algorithm}
\usepackage{algorithmic}
\usepackage{xcolor}
\usepackage{cite}

\title{
Reasoning with Scene Graphs for Robot Planning \\under Partial Observability
}

\author{Saeid Amiri, Kishan Chandan, and Shiqi Zhang

\thanks{Manuscript received: September, 09, 2021; Revised January, 01, 2022; Accepted January, 30, 2022.}
\thanks{This paper was recommended for publication by Editor Hanna Kurniawati upon evaluation of the Associate Editor and Reviewers' comments.
This work was supported by NSF (NRI-1925044), Ford, OPPO, and SUNY RF)}
\thanks{All authors are with the Department of Computer Science,
        the State University of New York (SUNY) at Binghamton, Binghamton, NY, 13902, USA. 
        Email: {\tt\small \{samiri1; kchanda2; zhangs\}@binghamton.edu}}%
        \thanks{Digital Object Identifier (DOI): see top of this page.}
}

\begin{document}

\maketitle



\begin{abstract}
Robot planning in partially observable domains is difficult, because a robot needs to estimate the current state and plan actions at the same time. 
When the domain includes many objects, reasoning about the objects and their relationships makes robot planning even more difficult. 
In this paper, we develop an algorithm called \emph{scene analysis for robot planning}~(SARP) that enables robots to reason with visual contextual information toward achieving long-term goals under uncertainty. 
SARP constructs scene graphs, a factored representation of objects and their relations, using images captured from different positions, and reasons with them to enable context-aware robot planning under partial observability. 
Experiments have been conducted using multiple 3D environments in simulation, and a dataset collected by a real robot. 
In comparison to standard robot planning and scene analysis methods, in a target search domain, SARP improves both efficiency and accuracy in task completion.
Supplementary material can be found at \url{https://tinyurl.com/sarp22}
\end{abstract}

\begin{IEEEkeywords}
Planning under Uncertainty, Probabilistic Inference, Semantic Scene Understanding.
 \end{IEEEkeywords}

\section{Introduction}

\IEEEPARstart{T}{here} has been great progress in development of service robots in the recent years, e.g.,~\cite{khandelwal2017bwibots,hawes2017strands}. 
Those robots are able to conduct everyday tasks in human-inhabited environments over extended periods of time. Robot perception in such domains is partial and unreliable, which brings a major challenge to robot decision making.

Partially Observable Markov Decision Process (POMDP) is a framework that models the uncertainty in both observations and action outcomes~\cite{kaelbling1998planning}, and has been used for policy generation in partially observable domains. 
However, the challenges are two-fold. First, constructing POMDPs requires that the robot has a complete world model, which tends to be infeasible in practice.
In particular, real-world environments (say a kitchen) frequently include many objects, making it troublesome to use POMDPs to have a universal representation of all objects. Second, the complexity of reasoning about these objects and their relationships grows exponentially as more objects are considered. 
In this paper, we aim to develop an approach that reasons with contextual information for scene analysis to enable POMDP-based robot planning.

One of the recent advancements in computer vision has been scene graph generation networks~\cite{zellers2018neural,li2017scene,chen2019knowledge,xu2017scene,rosinol20203d,engelcke2019genesis,lin2020space}. 
Given an image, scene graph systems generate a graph consisting of detected objects (e.g, a book and a table), their corresponding bounding boxes, and the relationships among the objects (e.g., \textit{book on a table}).
Scene graphs provide a robot with a structured understanding of the world in terms of objects, and their relations.
From the robotics perspective, however, current scene graph research has the limitation that the context analysis does not go beyond individual images, even though a robot can easily capture images from different angles and locations for analysis purposes. 
With the \emph{active} perception capabilities of robots, we have the objective of developing an approach for domain-wide active scene analysis for mobile robots. 

\begin{figure*}[t]
  \begin{center}
    \includegraphics[width=0.9\textwidth]{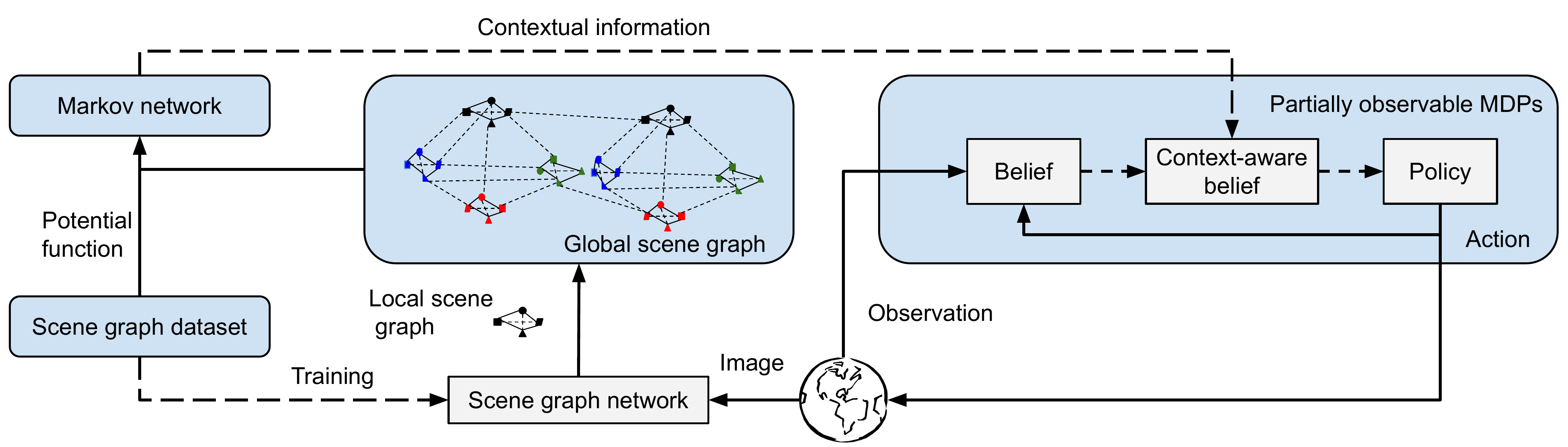}
    \caption{An overview of SARP, where the robot takes an action using the policy, and receives an observation from the world, updating the belief using the action and observation. In each timestep, a graph is accumulated using the perception sensor to bias the belief. This biased belief provides the contextual information to the policy. Top dashed lines show that the belief biasing not occurring in every iteration. The bottom dashed line indicates that the scene graph network is trained offline.   
    }
    \label{fig:overview}
  \end{center}
\end{figure*}

In this work, we develop an algorithm called \emph{scene analysis for robot planning} (\textbf{SARP}) for planning robot actions for context-aware, object-centric scene analysis. 
SARP uses local scene graphs of single images to build and augment global scene graphs toward context-aware robot planning under partial observability. 
An overview of SARP is shown in Figure~\ref{fig:overview}.
More specifically, a global scene graph is incrementally constructed ``on the fly'' using local scene graphs generated at different locations when new objects are perceived. 
Reasoning with this global scene graph produces useful information to help the robot estimate the current world state. 
This enhanced state estimation enables the robot to improve its performance in goal achievement. 

We have evaluated SARP using target search tasks where a robot needs to locate an object in an indoor environment. 
We use POMDPs to model the robot's perception and actuation skills~\cite{kaelbling1998planning}, use Neural Motifs~\cite{zellers2018neural} to compute local scene graphs, and use approximate inference methods to build Markov networks computed from large datasets.  
We have extensively evaluated SARP through comparisons with competitive baselines in simulation. 
Results show that SARP reduced the overall action costs by $16\%$ compared with a predefined action policy. 
Also, SARP helps the robot maintain its policy quality in the presence of an increased number of objects, and enables the robot to focus on the areas that are most relevant to the current task.

\section{Related Work}

This work aims to enable a robot to represent and reason with contextual information to guide robot planning under uncertainty. 
Researchers have developed algorithms that reason with contextual knowledge to guide sequential decision making~\cite{zhang2020survey}.
The contextual knowledge can be in a variety of forms, such as commonsense knowledge~\cite{davis2015commonsense}, action knowledge~\cite{nau2004automated,haslum2019introduction}, and graph-based knowledge~\cite{koller2009probabilistic,xu2017scene}.
Such contextual knowledge can also be leveraged to guide RL agents. 
In this section, we examine each of these categories.

\paragraph{Rule-based human commonsense knowledge}
Researchers have used rule-based commonsense knowledge to guide the robot planning under uncertainty~\cite{zhang2015corpp,lu2017leveraging,chitnis2018integrating}. 
In their methods, a robot reasons about human knowledge to compute an informative prior to help a probabilistic planner estimate the current world state, enabling the robot to achieve complex goals with less information-gathering behaviors. 
Our previous work added a learning component to further improve a robot's reasoning  and planning capabilities from its task completion experience~\cite{amiri2020learning}. 
Another example is an algorithm that uses first-order logic to construct decision tree policies for goal-oriented factored POMDPs~\cite{juba2016integrated}. 
Others have used action knowledge to build a hierarchical robot planner where the higher level computes a sequence of abstract actions and the lower level implements the higher-level actions using primitive behaviors~\cite{sridharan2019reba}.
Algorithm iCORPP enables a robot to reason with contextual knowledge to compute parameters of a planning agent's reward and transition functions~\cite{zhang2017dynamically}. 
Others have used commonsense knowledge to guide a classical planner to reason and plan in open worlds~\cite{jiang2019open,cui2021semantic}. 
Compared to those methods, SARP uses contextual, object centric information, in the form of a graph, to estimate the current world state and guide robot planning under uncertainty.  

\paragraph{Hierarchical frameworks}
Researchers have been using hierarchies to construct their framework's knowledge-base or planner~\cite{sridharan2019reba, zhu2020hierarchical}. 
In a recent work, researchers used a visual hierarchical planning algorithm for long-horizon manipulation tasks. Their framework integrates  neuro-symbolic task planning and graph-
based motion generation on graph-based scene representations~\cite{zhu2020hierarchical}. 
Their method has two-level abstractions of a manipulation scene with geometric scene graphs and symbolic scene graphs.
To enable a robot to operate in open-world domains, researchers have developed a three-layer hierarchy for reasoning about action knowledge and default knowledge~\cite{hanheide2017robot}. 
In particular, the knowledge at a higher level was used for correcting lower-level knowledge to guide probabilistic planning.
In comparison, SARP (ours) incrementally builds and reasons about scene graphs to guide robot planning under uncertainty. 

\paragraph{Graph-based human knowledge} 
Graph-based representations have been used for reasoning with contextual information to guide robot planning~\cite{zhu2020hierarchical,zeng2020semantic,wu2019bayesian}. 
For instance, researchers have used Conditional Random Fields~\cite{wallach2004conditional} to construct contextual knowledge bases for robot target search by maintaining a belief over the locations of target objects and landmark objects while exploiting the knowledge of their co-appearances~\cite{zeng2020semantic}. 
 
Other methods were developed to extract prior knowledge from data, e.g., using Long Short-Term Memory (LSTM) networks~\cite{hochreiter1997long}, to guide a planning agent in navigating new environments~\cite{wu2019bayesian}. 
In comparison to those methods, SARP leverages contextual information in the form of automatically constructed scene graphs, avoiding domain experts manually developing knowledge bases. 

A very recent work leverages scene graph reasoning to disambiguate human utterances when referring to various instances of the same object~\cite{yi2022incremental}.
Their framework reasons about static domains, whereas our robot leverages scene analysis to plan towards achieving long-term goals.

\paragraph{Knowledge-based RL}
RL agents are able to leverage contextual information as well. 
For instance, existing research has shown that an RL agent is able to reason with action knowledge to decompose complex tasks into smaller, tractable subtasks~\cite{yang2018peorl}. 
The concept of reward machines has been introduced for using temporal knowledge toward reusing interaction experience and creating extra feedback for learning purposes~\cite{icarte2018using}. 
Recent research has shown that action knowledge can be used to guide a model-based RL agent by providing optimistic, artificial interaction experience to speed up the learning process~\cite{hayamizu2021guiding,DBLP:journals/ai/LeonettiIS16}. 
We assume the availability of the world dynamics, and use planning under uncertainty methods (instead of RL) for robot decision making. 
To the best of our knowledge, SARP is the first that enables a planning agent to automatically construct, and use scene graphs for context analysis under partial observability. 

\paragraph{POMDP-based target search}
Several reasearchers have used POMDPs for the target search task~\cite{zheng2021spatial,wandzel2019multi,li2016act,xiao2019online}. 
In one work, a robot arm is tasked to search for an object in the clutter where it should learn the synergies of acting and seeing objects.
In this work, the policy learned by a POMDP model could determine when is a good time to look at and detect objects and when it should move objects around in order to reduce the clutter~\cite{li2016act}. 
In another work, a robot searched for multiple objects by specifiying their locations from the user query (e.g., \textit{Find the mugs in the kitchen and books in the library})~\cite{wandzel2019multi}.
In their framework, a robot can associate the locations to each object class so as to improve its search.
To the best of our knowledge, none of these works leverage graph reasoning for target search tasks.

\section{Preliminaries}
\label{sec:pre}

In this section, we describe the three building blocks of this research, namely scene graphs, and partially observable Markov decision processes.

\subsection{Scene Graphs}
\label{sec:sg}

Scene graph $G$ is a representation of the semantic content of an image, consisting of a set of bounding boxes $B = \{b_1, . . . , b_n\}$, a set of objects $V = \{o_0, o_1, \cdots \}$, and a set of binary predicates $E = \{r_0, r_1, \cdots \}$~\cite{johnson2015image}, 
A triplet of object-predicate-object is called a relationship. 
Given an image $I$, the probability distribution of the scene graph $pr(G|I)$ is decomposed into three components:
\begin{equation} \label{eqn:scene_graph}
    pr(G|I) = pr(B|I)pr(V|B, I)pr(E|V, B, I)
\end{equation}
where the bounding box component $pr(B|I)$ generates a set of candidate regions from the input image, $pr(V|B, I)$ predicts the class label for each predicted region, and $pr(E|V, B, I)$ predicts the predicate among objects, which is conditioned on the predicted labels.

\subsection{Partially Observable MDPs}

Markov decision processes (MDPs) can be used for sequential decision making under full observability~\cite{puterman2014markov}. 
Partially observable MDPs (POMDPs) generalize MDPs by assuming current state being partially observable~\cite{kaelbling1998planning}. 
A POMDP is represented as a tuple $(S, A, T, R, Z, O, \gamma)$ where $S$ is the state-space, $A$ is the action set, $T$ is the state-transition function, $R$ is the reward function, $Z$ is the observation set, $O$ is the observation function, and $\gamma$ is a discount factor that determines the planning horizon.

A robot maintains a belief state distribution $b$ based on observations ($z \in Z$) using the Bayes update rule: 
\begin{equation} \label{eqn:bayes}
    b'(s') = \frac{O(s', a, z)\sum_{s \in S} T (s,a,s')b(s)}{pr(z|a,b)}
\end{equation}
where $s$ is a state, $a$ is an action, $pr(z|a,b)$ is a normalizer, and $z$ is an observation. 
Solving a POMDP produces a policy that maps the current belief state distribution to an action toward maximizing long-term utilities. 

In this research, scene graphs and POMDPs are used for representing and reasoning about objects (and their relationships), and planning actions toward achieving long-term goals. 
The main contribution of this paper is the novel interplay between scene graphs and POMDPs for context-aware robot planning under partial observability. 
 
\subsection{Markov Networks}    
A Markov network is a graph that consists of variables (nodes) $\mathcal{X} =\{X_1,\cdots,X_n \}$ and undirected edges connecting pairs of nodes $(X_i,X_j)$~\cite{koller2009probabilistic}. 
Each edge is parameterized using a potential function (factor) $\phi$ that captures the affinities between the variables. 
The joint distribution of all variables are:
\begin{equation} \label{eqn:markov}
P(X_1, \cdots, X_n) = \frac{1}{Z}\prod_{i=1}^{m} \phi (D_{i})
\end{equation}
where $D_i$ is the $i$th edge of the network. A potential function of the edge connecting nodes $X_i$ and $X_j$, consists of $|X_i||X_j|$ values.

In the next section, we describe the algorithmic contribution of this work.

\section{Algorithm}

\paragraph{Problem Formulation}
In this work, we are interested in the problem of target search. 
A mobile robot receives the task of searching for object $Q$ and can navigate in the environment $E$ in order to find the object. 
The robot is provided with the environment map and is localized initially. 
Once the robot navigates in the environment sufficiently, it reports the location of the target object.
In order for the robot to better reason about $Q$, robot requires the scene graph network that is pretrained on a scene graph dataset $\mathcal{D}$.  
\paragraph{POMDP model}
In order to solve this target search task, we first define the POMDP model $\mathcal{M}$ as the tuple $<S,A,T,R,O,Z,\gamma>$. 
Its factored state space set $S$, is a Cartesian product of two dimensions, and the terminal state. 
$S^E$ includes a set of discrete partially observable locations of $Q$ (the target object) and $S^R$ includes the set of robot's fully observable locations. 
We define these equidistant, discrete locations manually in the robot's motion planning workspace. 
SARP's action set $A$ consists of navigation and termination actions. 
The robot can take navigation action $go_i \in A$ to go to the location $i$. 
We model the action transition function $T(s'|s,a)$ so that the robot can only go to its closest neighboring locations in the absence of obstacles. 
Each navigation action has a cost (negative value, $R(s,a)<0$), which is proportional to the distance the robot needs to travel to reach that location. 
The robot can take the termination action, and receive bonus (penalty) if it correctly locates (not find) the target object. 
Observation set $Z$ is $\{ $\textit{Detected}, \textit{NotDetected}, \textit{NotApplicable}$\}$ where \textit{Detected} happens when the perception detects the query object, \textit{NotDetected} happens when the query object is not detected, and \textit{NotApplicable} when the agent takes the termination action.     
To maximize its planning horizon, we set $\gamma$ to $0.99$.
SARP assigns the observation function $O= pr(z|s,a)$ based on the target's object detection accuracy on a test set of $\mathcal{D}$. 

\paragraph{Algorithm Description}
After defining the problem and the POMDP model, we present our novel algorithm, called \emph{scene analysis for robot planning} (SARP), for context-aware robot planning under partial observability. 
SARP computes a policy that enables a robot to accomplish its task using less action cost using the contextual knowledge. 
SARP is an object-centric algorithm that bridges the representation gap between visual scene analysis and robot planning under partial observability. 

Algorithm~\ref{alg:main} presents SARP, whose input is an object of interest that a mobile robot needs to locate ($Q$). 
SARP requires a POMDP solver for policy generation, a pre-trained scene graph generation network, a domain map for navigation, and a dataset $\mathcal{D}$ that consists of scene graphs. 
In Line~\ref{line:init_pomdp}, SARP constructs a POMDP according to the object of interest ($Q$), and computes policy $\pi$ using the provided POMDP solver. 
After that, there are a few steps for initializing beliefs (over object locations), and a scene graph (Lines~\ref{line:init_belief}-\ref{line:init_graph}). 
It should be noted that we maintain two beliefs $b$ and $b'$ over the location of the target object, where $b$ is updated using POMDP observations, and $b'$ is updated using the current $b$ and available contextual information. 
This design allows the robot to use different beliefs for decision making and action selection, and avoids possible issues caused by error propagation. Also, this mechanism enables a SARP agent to avoid reusing contextual information in belief updates.
This mechanism enables a SARP agent to avoid reusing contextual information in belief updates.

Lines~\ref{line:while_init}-\ref{line:while_end} form the main control loop of SARP, and it terminates when the current state is a terminal state. 
In each iteration, the robot uses a captured image to generate $G^L$, a local scene graph (Line~\ref{line:local}), and uses this local scene graph to update $G^G$, the global scene graph (Line~\ref{line:global}).

After that, SARP computes a potential function $\phi$ (Line~\ref{line:potential}) values for each edge, using the function $CALC \phi$~(Algorithm~\ref{alg:calc}). 
This function calculates four values for each relation $(v,\epsilon, v')$ in $G^L$ that serve as the potential function.
$CALC \phi$ queries the dataset $\mathcal{D}$, to find out the ratio of the times that each object ($v,v'$ or both) in a relation has appeared (not appeared) in $\mathcal{D}$, resulting in four values to form the potential function. 

We use $\phi$ to form a Markov network together with global scene graph $G$ in Line~\ref{line:markov} where SARP queries the number of times that each pair of nodes of an edge in $G^G$ has  appeared (not appeared) in $\mathcal{D}$ scene graphs.

In our implementation, $\mathcal{N}$ is incrementally updated in each iteration, if the robot detects new scene graphs (that were not previously detected), it will add the new nodes and relationships to the existing global scene graph $G^G$. We use the robot's localization and the camera depth sensor to approximately localize the detected objects on the map, in order to distinguish different instances of the same objects.

Lines~\ref{line:action}-\ref{line:bias_end} correspond to the belief update, and action selection processes of POMDPs. 
Contextual knowledge in the form of a Markov network ($\mathcal{N}$) is used for biasing belief $b$ only if the object of interest is visually detected in the current image. 
This is because of avoiding the bias drift that could result from too many biasing at every timestep.  
We use a \emph{belief propagation}\footnote{Any approximate inference method can be used.} method~\cite{yedidia2000generalized} to compute the probability of $Q$ being collocated with objects $V$ (Line~\ref{line:propagate}). 

\begin{equation} \label{eqn:BP}
    pr(Q|V) = BeliefPropagation(\mathcal{N}, Q, V)
\end{equation} 
where the computations of $pr(Q|V)$ and $b$ are independent. 
In Line~\ref{line:update}, SARP uses $pr(Q|G)$ to compute $b'$, the posterior belief distribution. 

\begin{algorithm}[t]
\small
\caption{SARP}
\textbf{Input:} Query object $Q$
\begin{algorithmic}[1]
\label{alg:main}
\REQUIRE a POMDP solver, a scene graph generation network, a domain occupancy grid map, robot localization software, scene graph dataset $\mathcal{D}$.

\STATE Construct a POMDP based on $Q$, and compute policy $\pi$\label{line:init_pomdp}
\label{line:init_start}
\STATE Uniformly initialize beliefs $b$ and $b'$ over $S^E$ \label{line:init_belief}
\STATE Initialize vertices $V=\{Q,Q'\}$ and edges $\mathcal{E}=\{ \epsilon_{Q-Q'} \}$, where $Q'$ is  a duplicate of $Q$ \label{line:init_vertices}
\STATE Initialize a scene graph: $G \leftarrow (V, \mathcal{E})$ \label{line:init_graph}
\WHILE{current state is not terminal} \label{line:while_init}
 
     \STATE Take an image, and generate local scene graph $G^L=(V', \mathcal{E'} )$, where $V'$ are detected evidence objects\label{line:local}
    \STATE  $V \leftarrow V \cup V';~\mathcal{E} \leftarrow \mathcal{E} \cup \mathcal{E'}$
    \STATE  $G^G \leftarrow (V, \mathcal{E} )$ \label{line:global}
    \STATE Compute potential function $\phi= CALC\, \phi(G^L,\mathcal{D})$\label{line:potential}
    \STATE Form a Markov network: $\mathcal{N} \leftarrow (G^G, \phi )$ \label{line:markov}
    \STATE Select action $a \gets \pi(b')$, and execute $a$ \label{line:action}
    \STATE Make observation $z$ on $Q$ \label{line:obs}
    \STATE Update $b$ based on $a$ and $z$\label{line:bayes_update}
    \IF{$z ~\textnormal{is}~ Detected$} \label{line:bias_begin}
        \STATE $pr(Q|V) \leftarrow BeliefPropagation(\mathcal{N}, Q, V)$ \label{line:propagate}
        \STATE $b' \leftarrow \eta \cdot  pr(Q|V) \cdot b$ \label{line:update}
    \ELSIF{$z ~\textnormal{is}~ NotDetected$}
        \STATE $b' \leftarrow b$
    \ENDIF \label{line:bias_end}
  \ENDWHILE \label{line:while_end}

\end{algorithmic}
\end{algorithm}

\paragraph{CALC function}

CALC function that takes as input the scene graph dataset $\mathcal{D}$ and the local scene graph $G^L$. 
For each pair of nodes $v$ and $v'$ that are connected in the $i$th edge $\epsilon$, the algorithm calculates four values by querying the number of times each node and their corresponding edge have (not) appeared in the dataset $\mathcal{D}$. 
We denote $v^1$ as the node exists in $\mathcal{D}$ and $v^0$ when it does not exist.
It returns the computed potential function $\phi$.

\begin{algorithm}[t]
\footnotesize
\caption{$CALC_\phi$ }
\textbf{Input:} Scene graph dataset $\mathcal{D}$, local scene graph $G^L$
\begin{algorithmic}[1]
\label{alg:calc}
\STATE Initialize the potential function set $\phi$ as empty
\FOR{$(v,\epsilon,v')^i$ in $G^L$ }
    \STATE $m \leftarrow $ number of images in $\mathcal{D}$ containing $v$ or $v'$.
    \STATE $\phi^i_{v^1,v'^1} \leftarrow$ (number of images in $\mathcal{D}$ containing $v,v', \textnormal{and}~\epsilon)/m $
    \STATE $\phi^i_{v^0,v'^0} \leftarrow $ (number of images in $\mathcal{D}$ containing $v,v', \textnormal{but not}~\epsilon)/m $
    \STATE $\phi^i_{v^0,v'^1} \leftarrow$ (number of images in $\mathcal{D}$ containing $v'$ without $v$)$/m$ 
    \STATE $\phi^i_{v^1,v'^0} \leftarrow$ (number of images in $\mathcal{D}$ containing $v$ without $v'$)$/m$
    \STATE $\phi \leftarrow \phi \cup \phi^i  $

\ENDFOR
\RETURN $\phi$
\end{algorithmic}
\end{algorithm}

SARP enables the agent to leverage contextual information towards task completion through building and reasoning with global scene graphs to guide a probabilistic planner. 
Next, we discuss our experiments for evaluating SARP.

\section{Experiments}
We conducted two sets of experiments in simulation where the robot is tasked with finding a target object accurately and as quickly as possible. In all the trials, the robot is provided with a domain map, a dataset of scene graphs, and a pretrained network for generating scene graphs. 
It receives 360-degree images of the environment as the input, where we leveraged our previous research on 360-degree robot vision~\cite{chandan2021learning}, and the output is the location of the target object. 
Our first baseline method is a naive POMDP planner~\cite{li2016act} (with uniform prior belief) where the robot action policy solely depends on the model of the world. 
The second baseline uses a predefined policy where the robot exhaustively visits all discrete positions, updates the belief at each timestep, and reports the object's location based on the argmax of the belief. 
Our third baseline method is CORPP~\cite{zhang2015corpp} as another competitive baseline where \textbf{only} the initial belief is biased based on the commonsense rules defined by the human developer in the form of logical probabilistic rules (e.g., \textit{a book is likely to be on a desk with $0.8$ probability}).
Similar to SARP, all the baselines maintain a belief of the object's location and update it at each timestep using Bayes update rule. However, none of them use graph reasoning like SARP.
We have two evaluation metrics. 
First, the average action cost that represent the average execution time of all actions taken until the terminal state. 
Second, is the average success rate in finding the target object's location correctly.  
By using SARP, we hypothesize that:
\begin{enumerate}
    \item The robot's overall action cost would be less compared to baselines (H-1). 
    \item SARP performs better than the baselines in action cost and success rate in domains with a large number of objects (H-2).
\end{enumerate}

The reward of successfully finding the target object is $100$, and the penalty of failure in finding the target object is $-100$. The reward for all actions $go_i$ is $-10$ which is proportional to the time it takes for the robot to execute the action. We solve the POMDP model using an off-the-shelf point-based system~\cite{kurniawati2008sarsop}.

We use Neural Motif \cite{zellers2018neural} for generating local scene graphs where there are a total of 50 predicate classes. 
The most prevalent objects in this dataset are humans and the most appeared predicates are \textit{in}, \textit{on}, and \textit{belongs}. 
Given an input image, Neural Motifs produces a scene graph which is a list of objects, their probabilities, relationships and bounding boxes.
SARP requires a dataset to assign the Markov network potential function (Line~\ref{line:potential} in Algorithm~\ref{alg:main}). We are using Visual Genome~\cite{krishna2017visual} that contains round 108K images, 3.8M objects, 2.8M relationships. 

\begin{figure}[t]
  \begin{center}
    \includegraphics[width=\columnwidth]{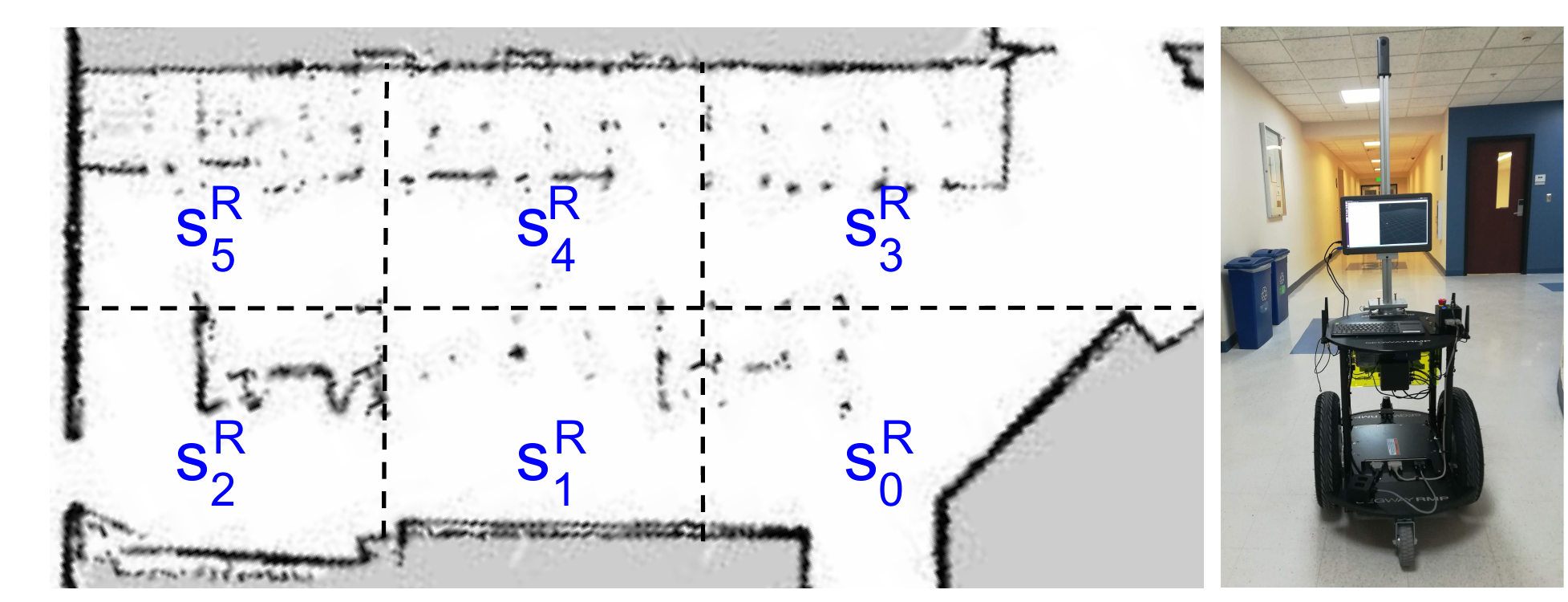}
    \caption{ (\textbf{Left}): Map of the environment where the robot can navigate. The discretized locations are shown in blue color. (\textbf{Right}): The Segway RMP110 mobile platform, equipped with 360-degree vision, used in this research. 
    }
    \label{fig:map}
  \end{center}
\end{figure}

\begin{figure*}[t]
  \begin{center}

    \includegraphics[width=0.8\textwidth]{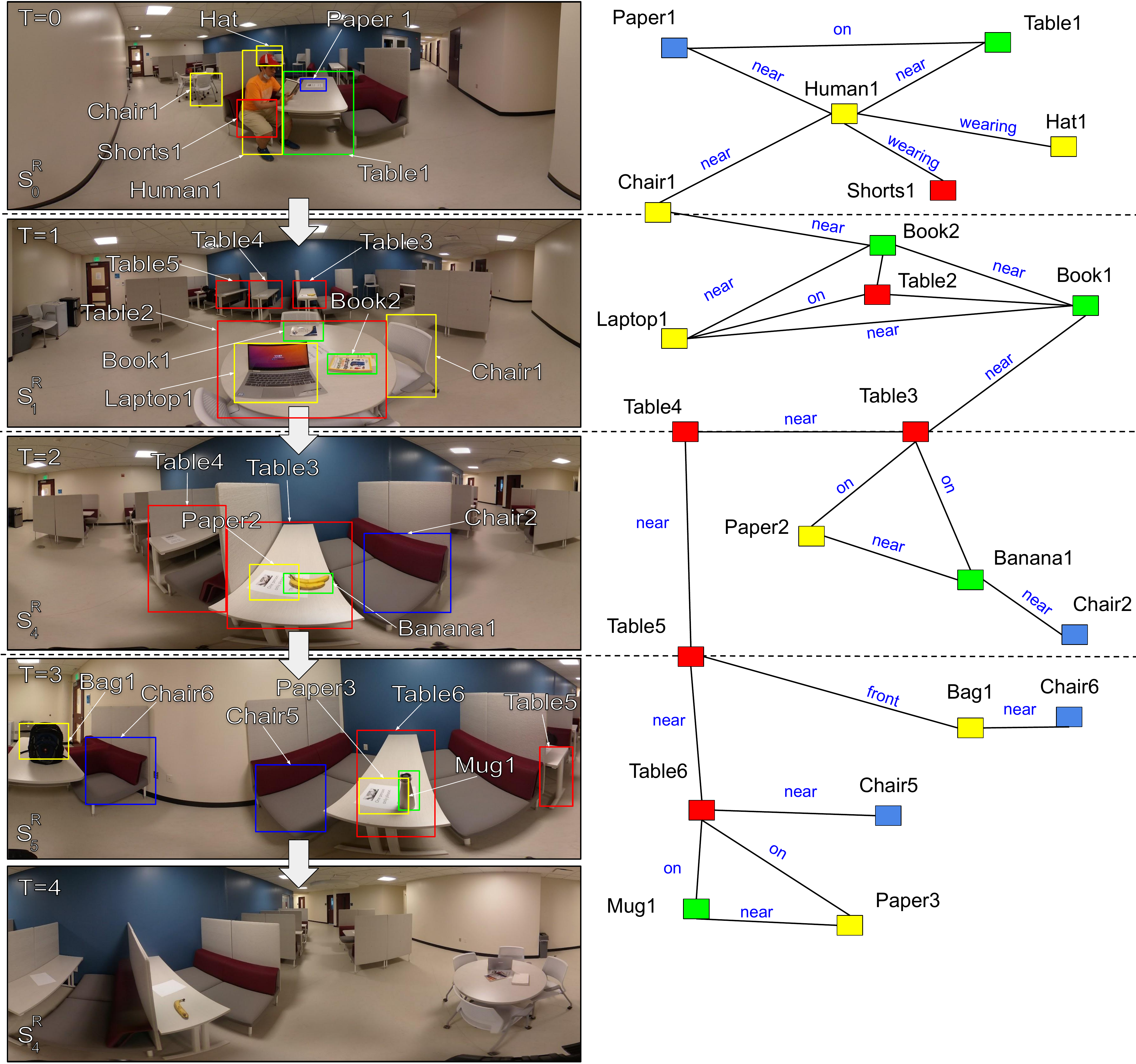}
    \caption{ The global scene graph being constructed incrementally as the robot navigates.
    Solid lines indicate the relationships between the detected objects, and dashed lines show how the global scene graph is completed from the local scene graphs.
    At timestep 4, the global scene graph is not augmented, since the robot has not detected new instances of objects.
    }
    \label{fig:gnn_demo_graph}
  \end{center}
\end{figure*}

\begin{table*}[ht]
\footnotesize
\caption{The experimental results from the hybrid and rendered datasets show that on average of 1500 trials, SARP performs with less overall action costs. Bold font shows the best result. The average number indicates the average number of objects involved in a single episode conducted in that environment. }
\begin{center}
\begin{tabular}{|c|p{1.5cm}|c|c|c|c|c|c|c|c|}
\hline
Dataset Type    &   Average \# of objects                              &  \multicolumn{2}{c|}{\textbf{SARP (Ours)}}       & \multicolumn{2}{c|}{CORPP}       & \multicolumn{2}{c|}{Uniform POMDP (Li et al.,2016)}  & \multicolumn{2}{c|}{Predefined}   \\
    \hline
    \multicolumn{2}{|c|}{}  &Cost (std.) & Success  &Cost (std.) &Success &Cost (std.) &Success &Cost (std.) &Success \\

    \hline
    \textbf{Hybrid}& 15 &\textbf{ 52.1(13.1)} & \textbf{0.89} & 82.5 (8.9) & 0.83 & 102.3 (14.6) &0.84 &61.1 (0.0) &0.7 \\
    \hline
    \hline
    \multicolumn{2}{|c|}{\textbf{Rendered}}&   \multicolumn{8}{c|}{}\\
    \hline
    Kitchen  & 70 & \textbf{41.7 (21.6) }  & 0.87   & 55.5 (18.4)   & 0.84 & 59.4 (24.1) &0.82 & 51.3 (0) & 0.79 \\
    \hline
    Living room  &37 & \textbf{43.5 (19.1) } & \textbf{0.91}    & 60.1 (33.8)   & 0.86 & 69.7 (23.9) & 0.73 & 57.8 (0) &0.79 \\
    \hline
    Bathroom  &  40   & 31.2 (16.4)  &  \textbf{0.75}   & 45.1 (17.6)     & 0.73  & 49.8 (21.7) & 0.71 & \textbf{21.2} (0)& 0.69 \\
    \hline
    Bedroom   & 35 & 23.9 (12.9)    & \textbf{0.74}    & 24.4 (19.1)   & 0.75  & 43.7 (22.5)& 0.68& \textbf{19.4} (0) & 0.64 \\
    \hline
    \hline
    Overall    & 45  & \textbf{35.7 (18.1)}    & \textbf{0.81}    & 46.3 (21.5)  & 0.79 &55.7 (23.0) & 0.73 &  37.3(0)& 0.72 \\
    \hline
\end{tabular}
\end{center}
\label{tab:hype_1}
\end{table*}

\subsection{Setup}
We used the robot to navigate through the environment and collect images in a hallway in an indoor educational environment (Figure~\ref{fig:map}) in order to build a dataset. We call it the \textit{hybrid} dataset. We used this dataset to simulate robot's behavior in the experiments. 
We manually placed multiple objects including \textit{banana, laptop, human, books, mug, etc.} at different locations.

\begin{figure}[t]
  \begin{center}

    \includegraphics[width=\columnwidth]{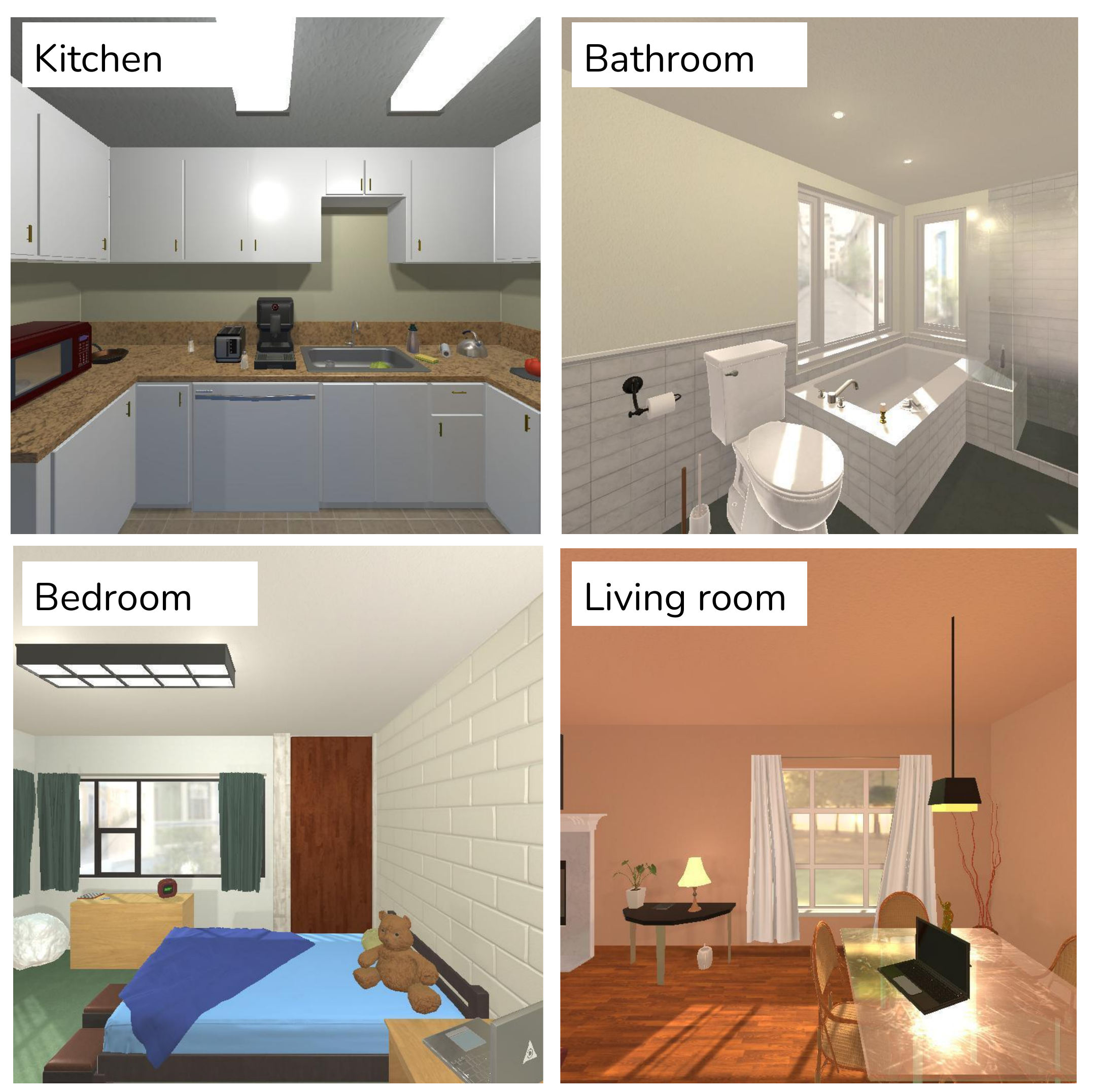}
    \caption{Different types of environment provided by AI2THOR including \textit{kitchen}, \textit{bathroom}, \textit{bedroom}, and \textit{living room}. }
    \label{fig:ai2thor_envs}
  \end{center}
\end{figure}

In addition to the dataset collected by the robot, we collected rendered images using an embedded agent in AI2THOR~\cite{kolve2017ai2}, an open-source interactive environment for embodied AI. 
AI2THOR provides 30 instances of four types of environments (shown in Figure~\ref{fig:ai2thor_envs}): \textit{Kitchens}, \textit{Living Rooms}, \textit{Bedrooms}, and \textit{Bathrooms}, totaling 120 environments where an embedded robot is able to take navigation actions. 
Navigation actions include moving and rotating in orthogonal directions. 
To make both datasets consistent, we manually create 360 images using the AI2THOR monocular camera. 
We use the default value of $1.5 m$ for the camera visibility and 90 degrees for the field of view. The actions executions are stochastic with the default Gaussian noise of average $0.001$ and standard deviation of $0.005$. 
There are a total of 125 objects in AI2THOR platform\footnote{https://ai2thor.allenai.org/ithor/documentation/objects/object-types/}, while for each scene in the platform, there are  61 number of objects on average.
We randomly select one of those objects for each individual trial in the experiments.
For inference on the scene graph, we use pgmpy library~\cite{ankan2015pgmpy}.

\begin{figure}[t]
  \begin{center}

    \includegraphics[width=\columnwidth]{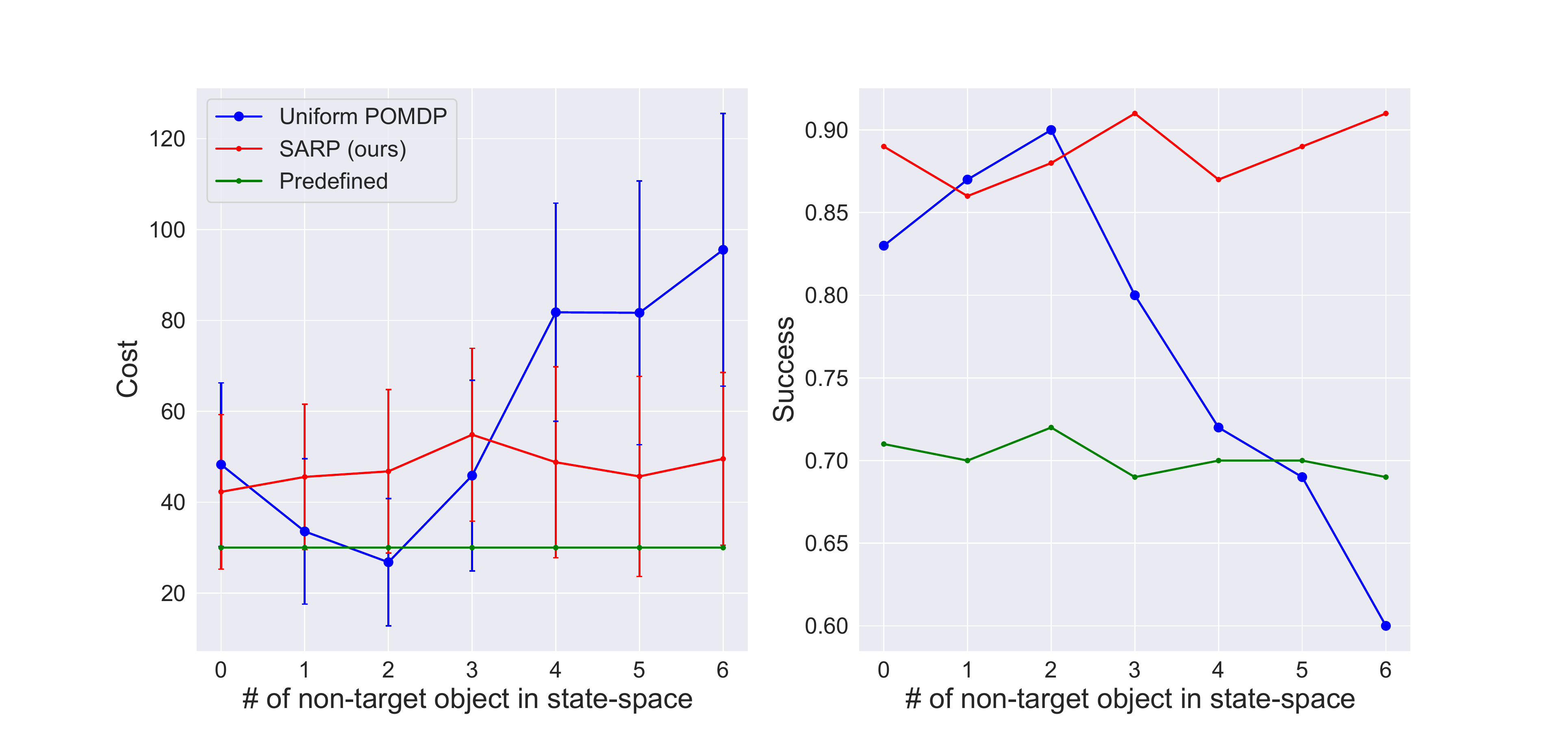}
    \caption{ Results of the experiment conducted on the renderred dataset. The state space consists of three locations, blue curve is the baseline that does not use contextual information but includes non-target objects in its state space incrementally, and the orange one is SARP that adds objects to the graph incrementally.}
    \label{fig:scalability}
  \end{center}
\end{figure}

\subsection{Results}

Table~\ref{tab:hype_1} shows the result of the first set of experiments that evaluates the first hypothesis (H-1) where we ran this experiment 1500 times over three batched of 500 experiments. We ran the experiment both using the dataset collected by the real robot and by AI2THOR.  
We call the first one, the \textit{hybrid} dataset and the latter one, the \textit{rendered} dataset. In the hybrid dataset, the average cost of target search for our robot is consistently less than all the baselines while all the methods maintain a high success rate.
In the rendered dataset, we categorize the results based on the four different types of indoor environments: \textit{kitchen}, \textit{living room}, \textit{bedroom} and \textit{bathroom}. 
We randomly selected five environment from each of the four types, totaling 20 different environments. 
In each environment, we conducted 100 target searches.
Except for the \textit{bedroom} and \textit{bathroom} environments where the predefined policy has lower cost, SARP produced the lowest average action cost while maintaining the highest accuracy consistently. The reason is that, in \textit{bedroom} and \textit{bathroom} environments, robot has a smaller navigation area. As a result, it takes less action costs using the predefined policy.    
To evaluate the second hypothesis (H-2), we incrementally added more objects to the POMDP baseline, and to the scene graph of SARP. 
Figure~\ref{fig:scalability}, shows that SARP's average cost remains almost the same with an average of 50.6 s while the POMDP baseline's cost increases as the number of additional objects is increased. 
With an increase in the number of non-target objects, the uniform POMDP baseline tries to take more actions to find the target object with more confidence, however with more than two non-target objects, it finds that taking more actions is not helpful any more, and therefore is not successful.
Also, the POMDP baseline failed to maintain the success rate at a high value due to its poor-quality policy.

\subsection{Enumerating objects of the same instance}
As the robot is navigating the environment, it may perceive different instances of the same objects. Our scene graph network is not able to enumerate these objects. To better differentiate these instances, we use a hashmap to store the labels and location(s) of the detected objects. This facilitates the situations where multiple instances of the same objects need to be distinguished. We approximate the detected objects’ locations using the centroid of the bounding boxes outputted by the scene graph. This provides the relative location of the object with respect to the robot. Then, we use coordinates transformation to get the object global scene graph.   
For the real robot experiments, we use the “Robot Operating System (ROS)~\cite{quigley2009ros} transforms” package to transform the locations to global coordinates. It should be noted that vision-based object association is generally difficult, and introduces errors into our system, which is beyond the scope of this paper.

\section{Demonstration}

\begin{center}
\footnotesize
\begin{table}
\caption{An example trial where the robot is tasked with the search of a \textit{banana} located in $s^E_4$ while robot is initially at $s^R_3$ .Here, we show how the belief is updated at each timestep.}
\label{tab:demo}

\begin{center}

\begin{tabular}{|c|c|c|} 

\hline
 & \textbf{A:} Action  & Robot's belief of the target   \\ 

Step & \textbf{O:} Observe &  [$s^E_0$ , $s^E_1$ ,$s^E_2$  ,     $s^E_3$  ,$s^E_4$,$s^E_5$ ]  \\ 
\hline

 &A: go $l1$ &    \\ 
 &O: No&   \\
0 &Update: & $[0.18 ,0.09 , 0.18     , 0.18   , 0.18,0.18 ]$ \\
 &Bias:No &   
\\
\hline
&A: go $l4$ &    \\ 
 &O: Yes &   \\
1 & Update: & $[0.15 ,0.07 , 0.15    , 0.15  , 0.33,0.15  ]$  \\
 & Bias:\textbf{Yes} & $[0.12 ,0.06 , 0.11     , 0.12  , 0.47,0.12  ]$  \\
\hline
 &A: go $l5$ &    \\ 
 &O: No&   \\
2 & Update: &$[0.13 , 0.05, 0.13   , 0.14   , 0.49,0.07  ]$   \\
 & Bias:No &   \\
\hline
&A: go $l4$ &    \\ 
 &O: Yes &   \\
3 & Update: & $[0.09 , 0.03 , 0.07 , 0.10  , 0.68,0.05  ]$   \\
 & Bias:\textbf{ Yes} & $[0.07, 0.01, 0.04    , 0.06   , 0.81,0.04  ]$  \\
\hline
4&A: terminate &    \\ 
\hline
\end{tabular}
\end{center}
\end{table}
\end{center}

In this demonstration trial, the robot was assigned the task of searching for a \textit{banana} as shown in Figure~\ref{fig:gnn_demo_graph} while its initial location is in location $0$. 
In Table~\ref{tab:demo}, we show an example trial where the robot is tasked with the search of a banana located at $s^E_4$.
As the robot takes action suggested by the policy, it updates its belief over possible locations of the target object. 
In this example, the robot follows the trajectory of $0 \rightarrow 1 \rightarrow 4 \rightarrow 5 \rightarrow 4$ until it terminates the trial and reports successfully that banana is in location $4$. 
The robot visits locations $0, 1, 4$, and $5$ to build and augment its scene graph. Figure~\ref{fig:gnn_demo_graph} shows the graph generated at each location. 
At timesteps $1$ and $3$, robot visits the location that it can easily detect the banana, therefore it biases its belief by inferring the whole scene graph. 
The overall cost for this trial is $60$.

\section{Conclusion \& Future work}
Probabilistic planning methods under partial observability allow the robot to accomplish complex tasks toward maximizing long-term goal. 
Scene graphs allow the detection of objects and their relationships in images.
Aiming at robots capable of accomplishing sophisticated tasks, we design a framework where a robot can use scene understanding information to obtain domain knowledge to provide contextual information to the robot planner where limited perception is a bottleneck.
Results show that, by using our approach, the robot can benefit from the contextual information in the form of graph network by reducing action collection cost and avoiding scalability issues. 

This work was based on a few assumptions that are sometimes unrealistic. The object detection accuracy and the set of are  all limited to the scene graph generation model quality. 
The scene graph we use, is incapable of face recognition, therefore we assume that humans are stationary (e.g, \textit{they are sitting at the time of the robot's task execution}).
In the future, we intend to further improve this work both in perception capabilities (e.g., \textit{enabling face detection}) and reasoning capabilities(e.g., \textit{considering cases where objects may be replaced during the task execution}). 

\section*{Acknowledgement}

This work has taken place at the Autonomous Intelligent Robotics (AIR)
Group, SUNY Binghamton. AIR research is supported in part by grants
from the National Science Foundation (NRI-1925044), Ford Motor Company
(URP Award 2019-2022), OPPO (Faculty Research Award 2020), and SUNY
Research Foundation.


{
\bibliographystyle{IEEEtran}
\bibliography{ref}
}

\end{document}